\documentclass{article}

\usepackage{booktabs}
\usepackage{color, colortbl}
\usepackage{graphicx}
\usepackage{microtype}
\usepackage{multirow}
\usepackage{pifont}
\usepackage{subcaption}
\usepackage{url}
\usepackage{wrapfig}
\usepackage{xcolor}
\usepackage[normalem]{ulem}

\usepackage{hyperref}

\usepackage[accepted]{icml2023}

\usepackage{amsmath}
\usepackage{amssymb}
\usepackage{mathtools}
\usepackage{amsthm}

\usepackage[capitalize,noabbrev]{cleveref}

\theoremstyle{plain}

\theoremstyle{definition}

\theoremstyle{remark}

\usepackage[textsize=tiny]{todonotes}

\usepackage{amsmath,amsfonts,bm}

\def\eqref#1{equation~\ref{#1}}

\def\1{\bm{1}}

\def\vone{{\bm{1}}}

\def\mA{{\bm{A}}}

\def\mD{{\bm{D}}}

\def\mW{{\bm{W}}}

\def\mY{{\bm{Y}}}

\DeclareMathAlphabet{\mathsfit}{\encodingdefault}{\sfdefault}{m}{sl}
\SetMathAlphabet{\mathsfit}{bold}{\encodingdefault}{\sfdefault}{bx}{n}

\def\sD{{\mathbb{D}}}

\DeclareMathOperator{\Tr}{Tr}

\newcommand{\cmark}{\ding{51}}%
\newcommand{\ourmethod}{\text{NOTELA}}
\newcommand{\pbf}{\mathbf{p}}
\newcommand{\xbf}{\mathbf{x}}
\newcommand{\ybf}{\mathbf{y}}
\newcommand{\thetabf}{\boldsymbol{\theta}}
\DeclareMathOperator{\precision}{Prec}
\DeclareMathOperator{\rank}{Rank}
\DeclareMathOperator{\lab}{Label}
\DeclareMathOperator{\map}{mAP}
\DeclareMathOperator{\cmap}{cmAP}
\definecolor{lightgreen}{rgb}{0.5, 0.93, 0.6}

\icmltitlerunning{In Search for a Generalizable Method for Source Free Domain Adaptation}

\begin{document}

\twocolumn[
\icmltitle{In Search for a Generalizable Method for Source Free Domain Adaptation}



\icmlsetsymbol{equal}{*}

\begin{icmlauthorlist}
\icmlauthor{Malik Boudiaf}{ets,atgoog}
\icmlauthor{Tom Denton}{goog}
\icmlauthor{Bart van Merri\"{e}nboer}{goog}
\icmlauthor{Vincent Dumoulin$^*$}{goog}
\icmlauthor{Eleni Triantafillou$^*$}{goog}
\end{icmlauthorlist}

\icmlaffiliation{ets}{\'{E}TS Montr\'{e}al}
\icmlaffiliation{atgoog}{Work done while interning at Google}
\icmlaffiliation{goog}{Google DeepMind}

\icmlcorrespondingauthor{Malik Boudiaf}{malik.boudiaf.1@etsmtl.net}

\icmlkeywords{Machine Learning, ICML, Source-Free Domain Adaptation}

\vskip 0.3in
]



\printAffiliationsAndNotice{\icmlEqualContribution} 

\begin{abstract}
Source-free domain adaptation (SFDA) is compelling because it allows adapting an off-the-shelf model to a new domain using only unlabelled data. In this work, we apply existing SFDA techniques to a challenging set of naturally-occurring distribution shifts in bioacoustics, which are very different from the ones commonly studied in computer vision. We find existing methods perform differently relative to each other than observed in vision benchmarks, and sometimes perform worse than no adaptation at all. We propose a new simple method which outperforms the existing methods on our new shifts while exhibiting strong performance on a range of vision datasets. Our findings suggest that existing SFDA methods are not as generalizable as previously thought and that considering diverse modalities can be a useful avenue for designing more robust models.
\end{abstract}

\section{Introduction}

Deep learning has made significant progress on a wide range of application areas. An important contributing factor has been the availability of increasingly larger datasets and models \citep{kaplan2020scaling,song2022clip}. However, a downside of this trend is that training state-of-the-art models has also become increasingly expensive. This is not only wasteful from an environmental perspective, but also makes the training of such models inaccessible to some practitioners due to the prohibitive resources required, or potential difficulties with data access. On the other hand, directly reusing already-trained models is often not desirable, as their performance can degrade significantly in the presence of distribution shifts during deployment \citep{geirhos2020shortcut}. Therefore, a fruitful avenue is designing \textit{adaptation methods} for pre-trained models to succeed on a new target domain, without requiring access to the original ({\em source}) training data (``source-free''). Preferably this adaptation can be performed \textit{unsupervised}. This is the problem of \emph{source-free domain adaptation} (SFDA) that we target in this work.

Several models have been proposed recently to tackle SFDA. However, we argue that \textit{evaluation} in this area is a significant challenge in and of itself: We desire SFDA methods that are \textit{general}, in that they can be used for different applications to adapt an application-appropriate pre-trained model to cope with a wide range of distribution shifts. Unfortunately, the standard SFDA evaluation protocols focus on a narrow set of shifts in vision tasks, leaving us with a limited view of the relative merits among different methods, as well as their generalizability. In this work, we address this limitation by studying a new set of distribution shifts. We expand on the existing evaluation methods, in order to gain as much new information as possible about SFDA methods. We also argue that we should target distribution shifts that are naturally-occurring as this maximizes the chances of the resulting research advances being directly translated into progress in real-world problems.

To that end, we propose to study a new set of distribution shifts in the \textit{audio} domain. We use a bird species classifier trained on a large dataset of bird song recordings as our pre-trained model. This dataset consists of \emph{focalized recordings}, where the song of the identified bird is at the foreground of the recording. Our goal is to adapt this model to a set of passive recordings (\emph{soundscapes}). The shift from focalized to soundscape recordings is substantial, as the recordings in the latter often feature much lower signal-to-noise ratio, several birds vocalizing at once, as well as significant distractors and environmental noise like rain or wind. In addition, the soundscapes we consider originate from different geographical locations, inducing extreme label shifts.

Our rationale for choosing to study these shifts is threefold. Firstly, they are challenging, as evidenced by the poor performance on soundscape datasets compared to focalized recordings observed by \citet{goeau2018overview,kahl2021birdnet}. Secondly, they are naturally occurring and any progress in addressing them can support ecological monitoring and biodiversity conservation efforts and research. Finally, our resulting evaluation framework is ``just different enough'' from the established one: It differs in terms of i) the modality (vision vs. audio), ii) the problem setup (single-label vs multi-label classification), and iii) the degree and complexity of shifts (we study extreme covariate shifts that co-occur with extreme label-space shifts). But it's not out of reach: SFDA methods designed for vision can be readily applied in our framework, since audio is often represented as spectrograms and thus can be treated as images.

We perform a thorough empirical investigation of established SFDA methods on our new shifts. Interestingly, the relative performance of established approaches varies significantly from observations made in common vision benchmarks, and in some cases we observe a {\em degradation} with respect to the pre-trained model's baseline performance. This striking finding leads us to explore simple modelling principles which we demonstrate result in consistently strong performance in the bioacoustics task considered and (importantly) in vision benchmarks as well. In the presence of extreme shifts, we observe that the confidence of the pre-trained model drops significantly (possibly also leading to miscalibration), which poses a challenge for entropy-based approaches. On the other hand, Noisy Student~\citep{xie2020self} and similar approaches are less sensitive to low model confidence but exhibit poor stability and require careful early-stopping---which is infeasible for SFDA because domain-specific validation data is unavailable. We hypothesize that leveraging the model's {\em feature space} as an additional ``source of truth'' helps stabilize adaptation, as this space carries rich information about the relationship between examples. In particular, we propose adding a Laplacian regularizer to Noisy Student, which we name NOisy student TEacher with Laplacian Adjustment (\ourmethod). 

Our contributions are: (i) we evaluate existing SFDA approaches on a challenging benchmark derived from a bioacoustics task and make observations on their generalizability which were not previously surfaced by common vision benchmarks; (ii) stemming from these observations, we advocate for the necessity of expanding the scope of SFDA evaluation in terms of modalities and distribution shifts; and (iii) we exemplify the benefits of this expanded scope by exploring simple modelling principles which, when combined, yield a more generalizable SFDA method.

\section{Related Work}
\label{sec:related_work}

See \autoref{tab:problem_settings} in the Appendix for a summary.

\textbf{Domain adaptation (DA).} DA assumes a setting in which labelled data is available for a source domain, and unlabelled data for a target domain. The goal is to maximize performance on the target domain. DA methods can be roughly divided into three types~\citep{sagawa2022extending}: \emph{domain-invariant training} (also called \emph{feature alignment}) aims to ensure that the features generated by the model for the source and target domain are indistinguishable by some metric~\citep{sun2016return,sun2016deep,tzeng2014deep,long2015learning,ganin2016domain,long2018conditional,tzeng2017adversarial,sankaranarayanan2018generate}; \emph{self-training} involves generating pseudo-labels for the unlabelled data~\citep{xie2020self}; and \emph{self-supervision} involves training an unsupervised/self-supervised model, later finetuned or jointly trained with supervision~\citep{shen2022how}.

{\bf Source-Free Domain Adaptation (SFDA) and Test-time Adaptation (TTA).} These methods additionally assume that the source data itself is not available, e.g., because of resource, privacy, or intellectual property concerns. The distinction between SFDA and TTA is subtle: the latter is transductive, motivated by an online setup where adaptation happens on (unlabelled) target examples as they appear and evaluation is subsequently performed on the same examples. SFDA considers an offline adaptation phase and the adapted model is then evaluated on held-out examples. In practice, though, the methods developed for either are similar enough to be applicable to both. Related problems also include black-box~\citep{zhang2021unsupervised}, online~\citep{yang2020casting}, continual~\citep{wang2022continual}, and universal~\citep{kundu2020universal} source-free domain adaptation.

Of the three types of DA methods discussed above, self-training most easily transfers to the SFDA and TTA settings~\citep{liang2020we,kim2021domain}, and we focus on this category since it's also the most generalizable to new modalities. Other methods use output prediction uncertainty for adaptation~\citep{yang2020casting,wang2021tent,roy2022uncertainty} or generative training to transform target domain examples or synthesize new ones~\citep{li2020model,hou2020source,kurmi2021domain,morerio2020generative,sahoo2020unsupervised}. Interestingly, \citet{boudiaf2022parameter} show that previous methods suffer from large hyperparameter sensitivity, and may \textit{degrade} the performance of the source model if not tuned in a scenario-specific manner; this violates the assumption that labelled target data is unavailable. 

{\bf Test-time Training (TTT).} TTT \citep{sun2020test} is a related problem where, like in TTA, a pre-trained model is adapted on the target test examples using a self-supervised loss, before making a prediction on those examples. Unlike SFDA and TTA, though, TTT modifies the source training phase to incorporate a similar self-supervised loss there too.

{\bf Domain generalization (DG).} In DG~\citep{wang2022generalizing}, like in SFDA, the target domain is unknown. However, unlike SFDA, no adaptation set is available. Instead the aim is to train a robust source model which works directly on new target distributions. Another important distinction is that DG assumes that information about the source domain is available during deployment on the target domain. A popular strategy for DG is to increase the source model's generalizablity by exposing it to diverse ``conditions'' at training time via domain randomization \citep{tobin2017domain} or adversarial data augmentation \citep{volpi2018generalizing,zhou2020deep}, or to learn domain-\textit{invariant} representations by training to match all available training ``environments'' \citep{arjovsky2019invariant,creager2021environment}, minimizing the worst-case loss over a set of such environments \citep{sagawa2020distributionally}, or decomposing the latent space or model weights into domain-specific and domain-general components \citep{ilse2020diva,khosla2012undoing}.

\section{Background}

\subsection{Problem formulation}
\label{sec:problem_formulation}
    
{\bf Notation.} In SFDA for classification, we assume access to a pre-trained model $f_{\thetabf}: \mathcal{X} \rightarrow \mathbb R^C$, where $\mathcal{X}$ denotes the input space and $C$ the number of classes. This model was trained on a source dataset $\sD_s$ sampled from a source distribution $p_s(\xbf)$, and needs to be adapted to a shifted target distribution $p_t(\xbf) \neq p_s(\xbf)$. We assume we only have access to unlabelled data $\sD_t^{\text{adapt}}$ sampled from $p_t$. The goal is to formulate an adaptation procedure $\mathcal{A}: (\thetabf_s, \sD_t^{\text{adapt}}) \rightarrow \thetabf_t$ that produces an adapted version of the original model using the unlabelled dataset. The adapted model's performance is then evaluated on held-out data $\sD_t^{\text{test}}$ sampled from $p_t$.

{\bf Single- vs. multi-label classification.} SFDA methods have traditionally addressed single-label classification, in which exactly one category of interest is present in a given sample. Multi-label classification relaxes this assumption by considering that any number of categories (or none) may be present in a given sample, which is common in real-world data. In the single-label case the output probability for sample $\xbf_i$ is noted $\pbf_{i,\thetabf} = \text{softmax}(f_{\thetabf}(\xbf_i)) \in [0,1]^C$. In the multi-label case, the predictions for each class are treated as separate binary classification problems, resulting in $\pbf_{i,\thetabf} = \left[ \sigma(f_{\thetabf}(\xbf_i)), 1 - \sigma(f_{\thetabf}(\xbf_i))\right]^\top \in [0, 1]^{2\times C}$, where $\sigma$ is the logistic function.
        
\subsection{Bioacoustics task}
\label{sec:bioacoustic_task}
    
We use {\em Xeno-Canto}~\citep[XC;][]{vellinga2015xeno} as the source dataset $\sD_s$ for bird species classification in the audio domain. XC is a large collection of user-contributed recordings of wild birds from across the world. Recordings are {\em focal} (targeted recordings of an individual captured in natural conditions, instead of the {\em passive} capture of all ambient sounds). Each recording is labeled with the species of the targeted individual; other birds may appear in the recording. 

To evaluate adaptation to distribution shifts, we use multiple collections of passive (also called {\em soundscape}) recordings from various geographical locations as our target datasets.
The soundscape datasets exhibit major covariate and label distribution shift from the source dataset. By virtue of being passively recorded, the annotated species are more occluded by environmental noise and distractors. Additionally, the geographical concentration of the datasets means that only a subset of XC's 
large number of species is present in each dataset, and the label distribution of those species does not necessarily follow that of XC. As a result, models trained on focal recordings have trouble generalizing to soundscapes recordings~\citep{goeau2018overview,kahl2021birdnet}.

\subsection{Vision tasks}

We evaluate on several vision robustness benchmarks, most of which are used by prior SFDA approaches: (i) CIFAR-10-C and ImageNet-C~\citep{hendrycks2019benchmarking}, a collection of corruptions applied to the CIFAR-10 and ImageNet test sets spanning 15 corruption types and 5 levels of severity; (ii) ImageNet-R~\citep{hendrycks2021many}, which consists of 30,000 images of 200 of ImageNet's classes obtained by querying for renditions such as ``art'', ``cartoon'', ``graffiti'', etc.; (iii) ImageNet-Sketch ~\citep{wang2019learning}, which consists of 50,000 images from querying Google Images for ``sketch of \{class\}" for all ImageNet classes; and (iv) VisDA-C~\citep{peng2017visda}, which contains images of 12 object classes spanning synthetic and real domains.

\subsection{Evaluation methodology}
        
Adaptation in SFDA is fully unsupervised and hence there is no annotated validation set for each target domain. This is a significant challenge for model selection and evaluation, as evidenced by recent work on domain generalization~\citep{gulrajani2021search}. In line with recommendations made by \citet{gulrajani2021search}, we disclose the model selection strategy used in our work, which for simplicity is shared across evaluated approaches. We hold out one domain for audio and one domain for vision which are used for validation and hyperparameter selection (details in \autoref{sec:results}). An extensive hyperparameter search is conducted for every approach. In line with the methodology prescribed by SFDA, we also partition the evaluation data for each distribution shift into the adaptation set $\sD_t^{\text{adapt}}$ (75\% of the data) and the test set $\sD_t^{\text{test}}$ (25\% of the data).

\subsection{Categorization of approaches}
\label{sec:categorization}
    
We will evaluate several methods which have been proposed for (source-free) domain adaptation on our new shifts. We choose to investigate methods that fit two criteria: generality across tasks and modalities, and strong performance. In this section we present our categorization of methods we consider, which we then build upon in Section \ref{sec:our_method}.

{\bf Entropy Minimization (EM).} These methods enforce the \textit{cluster} assumption, i.e., that the boundaries described by the model's head should not cross any high-density region of samples in the feature space. Geometrically, this enforces large margins between the classifier's boundaries and the provided samples by encouraging the model to output increasingly confident predictions for the unlabelled samples \citep{grandvalet2004semi}. 

We evaluate {\bf TENT}~\citep{wang2021tent} as a representative example of the EM approach. TENT adapts the source model by minimizing the entropy of its predictions through tuning the normalization layers' channel-wise scaling and shifting parameters, and updating the layers' population statistics estimates accordingly. 

{\bf Teacher-Student (TS).} This is a self-training paradigm where a teacher provides \emph{pseudo-labels} for the unlabelled examples, and a student is then trained to predict them. More formally, TS minimizes
\begin{equation}
\label{eq:TS_objective}
\begin{split}
    \min_{\ybf_{i:N}, \thetabf}~ \Tr \left(- \frac{1}{N} \sum_{i=1}^N \ybf_i^\top \log\left(\pbf_{i,\tau(\thetabf)}\right)\right) & \\
    \text{s.t} \quad \mathbf{1}^\top\ybf_i=\mathbf{1}, ~ \ybf_i \geq 0 ,&
\end{split}
\end{equation}
where $\ybf_i$ and $\pbf_{i,\tau(\thetabf)}$ represent the pseudo-label and the model's soft predictions for the sample $\xbf_i$.\footnote{Recall from Section~\ref{sec:problem_formulation} that in the multi-label case we consider $C$ separate binary classification problems, and hence $\ybf_i$ and $\pbf_{i,\tau(\thetabf)}$ are $2 \times C$ matrices. The trace in Equation~\ref{eq:TS_objective} sums the objectives for each of these $C$ problems.} The pseudo-label is dependent on the model's weight's, $\thetabf$, which are transformed using a weight transformation, $\tau$, which is typically set to the identity (which we denote $\text{Id}$) in TS methods.

Optimization happens in an alternating fashion between student and teacher updates: The \emph{teacher-step} minimizes \eqref{eq:TS_objective} w.r.t. the pseudo-labels $\{\ybf_i, \dots, \ybf_N\}$ while the \emph{student-step} minimizes \eqref{eq:TS_objective} w.r.t. $\thetabf$ using gradient descent. Different TS methods utilize soft~\citep{xie2020self} or hard~\citep{lee2013pseudo} pseudo-labels.

Intuitively, TS can be seen as an indirect way of minimizing entropy: By training to predict each example's pseudo-label (i.e., its \textit{most likely} label, based on its confidence), the model reinforces its own predictions, thereby increasing its confidence. At the same time, though, it has a \emph{consistency maximization} flavour: Because of the alternation between teacher and student updates, predicting the correct pseudo-label requires consistency throughout time.

We consider three methods from this category in our investigation: First, {\bf pseudo-labelling}~\citep[PL;][]{lee2013pseudo} assigns pseudo-labels to unlabelled examples by picking the maximum-probability class according to the trained model. In the context of SFDA, this translates into assigning pseudo-labels to unlabelled target domain examples using the model trained on the source dataset. Second, we consider an adaptation of {\bf DUST}~\citep{khurana2021unsupervised}, originally developed for DA (instead of SFDA), for automated speech recognition. DUST is also based on pseudo-labels, but only uses \emph{reliable} samples, as measured by the consistency of the predictions obtained by performing multiple stochastic forward passes through the model. Finally, {\bf SHOT}~\citep{liang2020we} adapts the feature extractor to the target domain while freezing the classifier head. It performs adaptation through a combination of nearest-centroid pseudo-labelling and an information maximization loss.

{\bf Denoising Teacher-Student (DTS).} This method builds upon the TS framework by adding some form of ``noise'' to the student, while keeping the teacher clean. Intuitively, predicting clean pseudo-labels from noisy student predictions leads to maximizing another type of consistency: between different views of the same inputs. Mathematically, DTS differs from TS by setting $\tau \ne \text{Id}$ during the student's forward pass, while keeping it set to the identity during the teacher's forward pass. 

Ideas related to DTS have been explored in semi-supervised \citep{xie2020self, miyato2018virtual} and self-supervised learning \citep{grill2020bootstrap, chen2020simple}. Notably, Noisy Student \citep{xie2020self} is a popular representative that we build upon. However, to keep the approach light, both in terms of computation and hyper-parameter load, we consider a simplified model in our investigation, where the same network is used for both the teacher and the student, and dropout is the sole source of noise. We refer to this variant as {\bf dropout student (DS)}. 

{\bf Manifold regularization (MR).} Manifold regularization exploits the assumption that target data forms well-defined clusters in the feature space of the source model, by explicitly enforcing the \textit{cluster assumption}. From this family of methods, we consider \textbf{NRC} \cite{yang2021exploiting}, which forces a target sample and its nearest-neighbors, as well as its `extended nearest-neighbors', to have similar predictions.

Our proposed model, presented in \autoref{sec:our_method}, combines elements of MR and DTS.

\begin{figure*}[t]
    \centering
    \includegraphics[width=\textwidth]{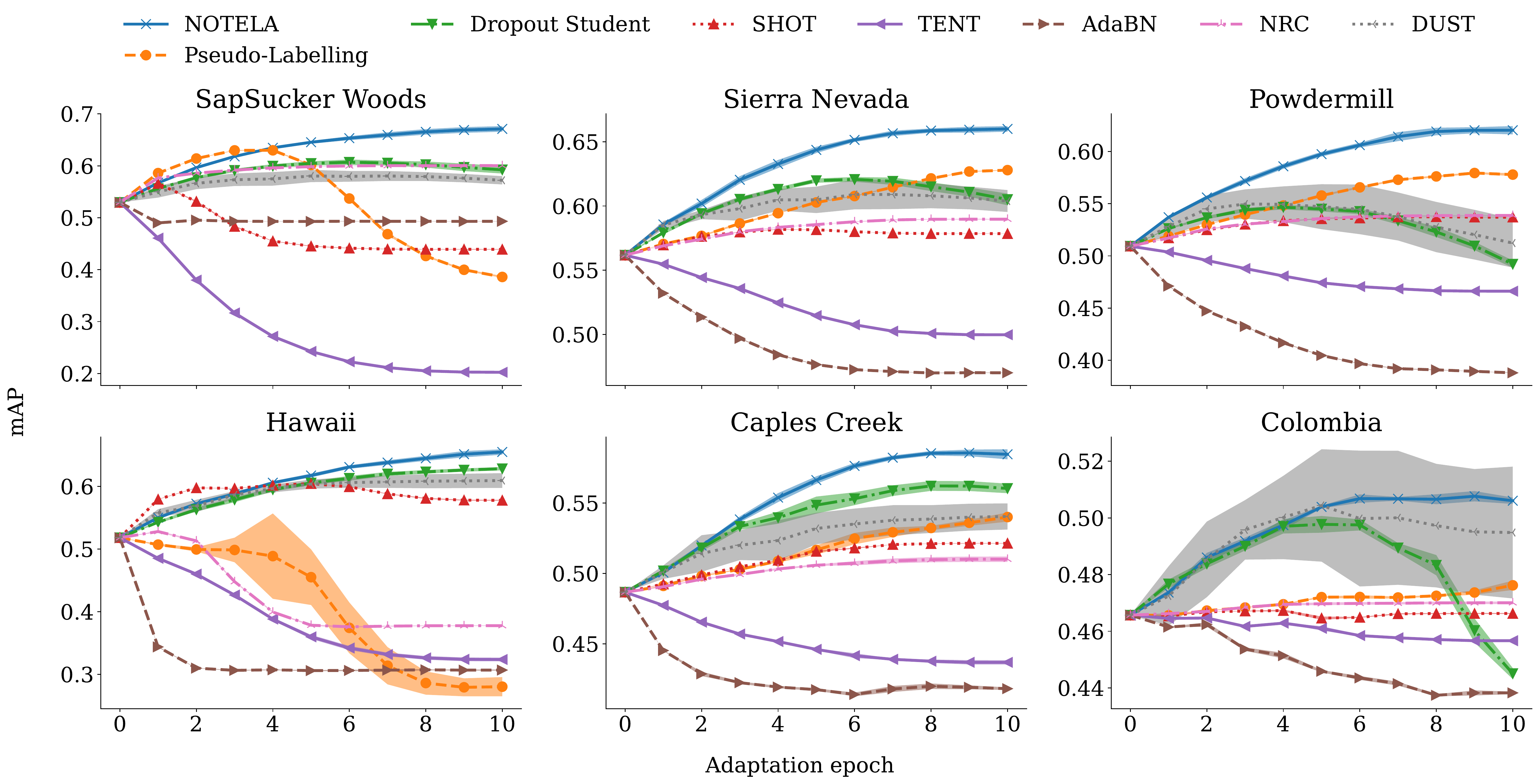}
    \caption{All but one SFDA method fail to consistently improve the source model in terms of $\map$ on $\sD_t^{test}$ across distribution shifts\protect. Dropout Student succeeds, but only through early stopping, which is infeasible in the SFDA setting. \ourmethod{}~achieves consistently stable convergence while improving upon Dropout Student's performance. Note that this analysis's purpose is to illustrate the failure modes of SFDA methods; those plots cannot be used for hyperparameter selection because we are looking at the test sets.}
    \label{fig:test_plots}
\end{figure*}

\section{Laplacian Adjustment}
\label{sec:our_method}

As we will demonstrate in~\autoref{sec:audio_results}, existing SFDA methods struggle to perform consistently well on bioacoustics distribution shifts. This motivates the exploration of a combination of simple modelling principles which we hypothesize will result in consistently strong performance on different datasets, problem settings and modalities. We name this combination NOisy student TEacher with Laplacian Adjustment (\textbf{\ourmethod{}}).

In our experiments, we observe that EM struggles in the face of severe distribution shifts resulting in models with very low confidence. We also note that TS consistently improves upon EM, signaling that consistency maximization is a useful ``auxiliary task'', perhaps due to being more robust to low model confidence. The fact that DTS---with the additional consistency maximization ``task'' between clean and noisy views---further improves performance in our experiments strengthens this hypothesis. Starting from DTS, we seek to further improve it by utilizing another source of information that we hypothesize is also robust to low model confidence. Specifically, while the model outputs may have low confidence, its \textit{feature space} may still carry useful information about the relationship between examples.

From this hypothesis, and drawing inspiration from classic ideas on manifold regularization~\citep{belkin2006manifold}, we suggest probing the feature space directly as an \textit{auxiliary source of truth}. \ourmethod{}~instantiates this idea by encouraging that nearby points in the feature space be assigned similar pseudo-labels, similar to NRC. We hypothesize this may also help with stability, since the targets that the student is asked to predict will vary less over time, due to the slower-changing similarity in feature space. NRC uses a sophisticated definition of neighbourhood which weighs reciprocal, non-reciprocal, and extended nearest neighbours differently. In contrast, \ourmethod~simplifies this by considering only reciprocal nearest-neighbours. NRC and \ourmethod~also differ by their class-marginal pseudo-label prior (uniform vs. no prior) and the loss function used (dot product vs. cross-entropy).

\textbf{Formalization.} We augment the denoising teacher-student formulation in \eqref{eq:TS_objective} with a Laplacian regularization:
\begin{equation}
\label{eq:laplacian_objective}
\begin{split}
    \min_{\ybf_{1:N}, \thetabf} \Tr\Bigg( \frac{1}{N} \sum_{i=1}^N \ybf_i^\top \Bigg[ &-\log\left(\pbf_{i,\tau(\thetabf)}\right) + \alpha \log(\ybf_i) \\
        & - \lambda \sum_{j=1}^N w_{ij} ~ \ybf_j\Bigg]\Bigg).
\end{split}
\end{equation}
There are two changes in \autoref{eq:laplacian_objective} compared to \autoref{eq:TS_objective}. First, we introduce a scalar weight $\alpha \in \mathbb R$ that controls the softness of pseudo-labels, which we treat as a hyperparameter. Second, we have added the third term that represents a Laplacian regularizer. The value $w_{ij}$ denotes the \textit{affinity} between samples $i$ and $j$ and is obtained by the penultimate layer of the network.

\textbf{Optimization.} Disregarding the pairwise Laplacian term allows to directly obtain a closed-form solution to \autoref{eq:laplacian_objective}, namely $\ybf_i \propto \pbf_i^{1/\alpha}$. However, adding the pairwise affinities makes optimization more challenging. We simplify the problem by linearizing the Laplacian term, which allows us to recover a closed-form solution:
\begin{align} \label{eq:laplacian_updates}
    \ybf_i \propto \pbf_i ^{1/\alpha} \odot \exp \left(\frac{\lambda}{\alpha} \sum_{j=1}^N w_{ij} \pbf_j \right)\,.
\end{align}
The full proof of \autoref{eq:laplacian_updates} can be found in the Appendix. Furthermore, we show in Appendix \ref{sec:laplacian_proof} that under the assumption of positive semi-definite affinity matrix $(w_{ij})$, \autoref{eq:laplacian_updates} becomes an instance of a concave-convex procedure \citep[CCP;][]{yuille2003concave}.

\textbf{Complexity.} Theoretically, the added Laplacian regularization scales quadratically in the number of samples $N$. In practice, we set $w_{ij}= w_{ji} = \frac{1}{d}$ where $0 < d \leq k$ is the number of mutual $k$-nearest neighbours~\citep{brito1997connectivity} of samples $i$ and $j$, and $w_{ij}= w_{ji} = 0$ if these samples are not in each other's $k$-nearest neighbours lists. Finding the $k$-nearest neighbours can be done with $\mathcal{O}(N\log N)$ average time complexity. 
\autoref{eq:laplacian_updates} scales as $\mathcal{O}(NCk)$, but can be fully vectorized across samples.

\begin{table*}[t]
    \caption{Test results on the 6 test target domains (averaged over 5 random seeds) using hyperparameters selected on the validation domain.}
    \label{tab:bioacoustics}
    \vskip 0.15in
    \begin{center}
    \footnotesize
    \resizebox{\textwidth}{!}{
        \begin{tabular}{lllllllllllll}
            \toprule
            \multirow{2}{*}{Method}  & \multicolumn{2}{c}{S. Nevada} & \multicolumn{2}{c}{Powdermill} & \multicolumn{2}{c}{Hawai'i} & \multicolumn{2}{c}{Caples} & \multicolumn{2}{c}{SSW} & \multicolumn{2}{c}{Colombia} \\
            \cmidrule(lr){2-3} \cmidrule(lr){4-5} \cmidrule(lr){6-7} \cmidrule(lr){8-9} \cmidrule(lr){10-11} \cmidrule(lr){12-13}
             & $\map$ & $\cmap$ & $\map$ & $\cmap$ & $\map$ & $\cmap$ & $\map$ & $\cmap$ & $\map$ & $\cmap$ & $\map$ & $\cmap$\\
            \midrule
            Source & 56.2 & 36.4 & 50.9 & 33.0  & 51.8 & 33.2  & 48.7 & 41.8  & 53.0 & 33.7  & 46.6 & 38.5  \\
            AdaBN  & 47.0 & 34.3 & 38.8 & 30.7  & 30.7 & 28.0  & 41.8 & 38.8  & 49.3 & 34.2  & 43.8  & 35.5  \\
            SHOT   & 57.8 & 37.8 & 53.7 & 33.1  & 57.7 & 29.1  & 52.1 & 43.7  & 43.9 & 24.9  & 46.6 & 39.3  \\
            TENT   & 50.0 & 35.1 & 46.6 & 32.7  & 32.4 & 25.7  & 43.7 & 39.6  & 20.2 & 14.0  & 45.7 & 37.6  \\
            PL     & 62.8 & 39.3 & 57.8 & 33.3  & 28.0 & 25.6  & 54.0 & 43.8  & 38.6 & 30.6  & 47.6 & 39.6  \\
            DS     & 60.5 & 37.2 & 49.2 & 33.3  & 62.8 & \bf 36.4  & 56.0 & 41.7 & 59.2 & 33.2  & 44.5 & 41.6  \\
            DUST   & 60.4 & 35.1 & 51.2 & 31.5  & 61.0 & 35.7  & 54.1 & 40.3  & 57.2 & 32.3  & 49.5 & 35.6  \\
            NRC    & 59.0 & 38.0 & 53.9 & 33.3  & 37.8 & 28.7  & 51.0 & 43.5  & 60.0 & 39.3  & 47.0 & 39.5  \\
            \rowcolor{lightgreen!25} \ourmethod{}  & \bf 66.0 & \bf 40.0 & \bf 62.0 & \bf 34.7 & \bf 65.5 & 36.2 & \bf 58.5 & \bf 44.7 & \bf 67.1 & \bf 42.7 & \bf 50.6 & \bf 44.7 \\
            \bottomrule
        \end{tabular}
    }
    \end{center}
\end{table*}

\begin{table*}[t]
    \caption{Test accuracy on the 6 test target domains in the \emph{single-label} scenario (averaged over 5 random seeds) using hyperparameters selected on the validation domain.}
    \label{tab:single_label_bioacoustics}
    \vskip 0.15in
    \begin{center}
    \footnotesize
    \begin{tabular}{lccccccc}
        \toprule
        Method  & S. Nevada & Powdermill & Hawai'i & Caples & SSW & Colombia \\
        \midrule
        Source & 29.9 & 51.4 & 24.0 & 15.4 & 33.5 & 19.6 \\
        AdaBN & 14.2 & 30.9 & 13.0 & 10.1 & 25.5 & 14.2 \\
        DUST & 49.8 & 62.9 & 44.2 & 34.8 & 43.2 & \bf 32.7 \\
        SHOT & 23.9 & 15.2 & 6.8 & 17.6 & 21.9 & 18.9 \\
        TENT & 7.3 & 74.3 & 29.9 & 8.0 & 5.4 & 5.0 \\
        PL & 60.2 & \bf 93.1 & 50.0 & 33.7 & 58.7 & 24.0 \\
        NRC & 42.8 & 55.2 & 26.9 & 27.5 & 16.7 & 29.4 \\
        DS & 62.5 & 44.6 & \bf 56.1 & 39.3 & 49.4 & 27.7 \\
        \rowcolor{lightgreen!25} \ourmethod{} & \bf 73.5 & 90.5 & 29.1 & \bf 43.9 &  \bf 63.3 & \bf 32.5 \\
        \bottomrule
    \end{tabular}
    \end{center}
\end{table*}

\section{Experiments}
\label{sec:results}

\subsection{Bioacoustics Task}
\label{sec:audio_results}
    
{\bf Data processing and source model.} XC is a large dataset containing a total of 10,932 species. We process XC recordings by resampling the audio to 32 kHz and extracting 6-second slices of relevant audio (see Appendix~\ref{app:xc_processing} for details). We process soundscape recordings by extracting 5-second slices using the provided bounding-box labels (see Appendix~\ref{app:soundscapes_processing} for details).

During training, we take a random 5-second crop and the audio signal's gain is normalized to a random value between 0.15 and 0.25. A technique similar to LEAF~\citep{zeghidour2021leaf} is used to convert the waveform into a spectrogram. Unlike LEAF, we do not learn a Gaussian lowpass filter and instead apply the Gabor filters with a stride of 320. The resulting output is a power-spectrogram with a time resolution of 100 Hz.

The most comprehensive bird species classifier we are aware of is BirdNet~\citep{kahl2019identifying,kahl2021birdnet}, however the publicly available checkpoints are trained with a combination of focal and soundscape recordings (including many of the test sets considered in this paper),\footnote{Personal correspondence with the authors.} which is incompatible with our SFDA methodology. We instead use an EfficientNet-B1~\citep{tan2019efficientnet} model we trained ourselves. The output of this model is flattened and projected into a 1280-dimensional embedding space. In addition to species prediction, the model is trained with three auxiliary losses for the bird's order, family, and genus (each having a 0.25 weight).

{\bf Metrics.} Unless otherwise specified, we use sample-wise mean average precision ($\map$) and class-wise mean average precision ($\cmap$) for evaluation. Both of these metrics are threshold-free and appropriate for multi-label scenarios. Each can be interpreted as a multi-label generalization of mean reciprocal rank, where ranking of model logits is performed either per-sample (for $\map$) or per-class (for $\cmap$). See \autoref{sec:metrics_equations} in the appendix for formal definitions.

\begin{figure}[t]
    \centering
    \includegraphics[width=0.95\linewidth]{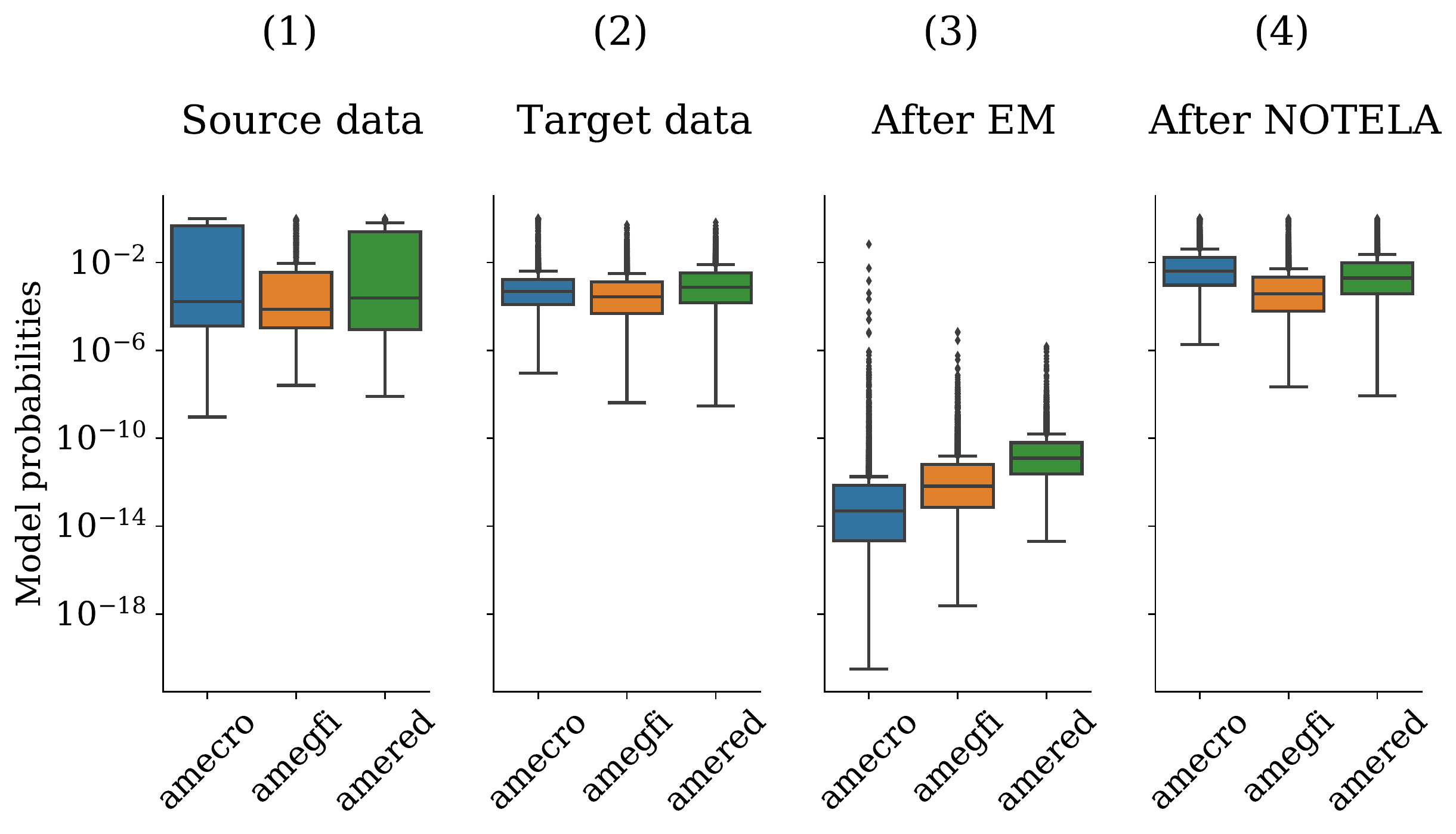}
    \caption{(1) and (2): When shifted to a new domain, the pre-trained model exhibits \emph{lower confidence} than on the original source data, as seen from the lower per-class probability distributions assigned by the model. (3): Applying entropy minimization only worsens the problem, and leads to \textit{general collapse}. (4): Our method does not suffer from the same phenomenon.
    Species names are represented using their corresponding eBird~\citep{sullivan2009ebird,sullivan2014ebird} species code.}
    \label{fig:model_calibration}
    \vspace{-10pt}
\end{figure}

The $\map$ metric measures the ability of the model to assign higher logits to any species present in an example.  By contrast, $\cmap$ is the mean of the model's per-species classification quality (similar to the average of per-species AUC scores).  Note that class-averaging in $\cmap$ corrects for class imbalance, while $\map$ reflects the natural data distribution. To avoid noisy measurements we only consider species with at least five vocalizations in the dataset when computing $\cmap$.

{\bf Baselines.} In addition to {\bf TENT}, {\bf pseudo-labelling (PL)}, {\bf SHOT}, {\bf dropout student (DS)}, {\bf DUST} and {\bf NRC} which we described in section \ref{sec:categorization}, we also consider {\bf AdaBN}~\citep{li2018adaptive}. This method recomputes the population statistics of batch normalization~\citep{ioffe2015batch} layers in the pre-trained model using the unlabelled target dataset.
        
{\bf Hyperparameter selection.} We use the High Sierras dataset as the held-out domain. For every method, we search over hyperparameters such as the learning rate and its schedule, the subset of parameters to adapt (all or only batch norm), and whether to use dropout during adaptation. Additionally, we search over specific hyperparameters such as the $\beta$ weight in SHOT, the confidence threshold in PL, or the weights $\{\alpha, \lambda\}$ in \ourmethod{}. The overview of tuned hyperparameters, resulting in $O(200)$ experiments per method, can be found in the appendix.

\begin{table}[t]
    \caption{Test results on the validation domain (High Sierras) using hyperparameters selected on that domain.}
    \vskip 0.15in
    \centering
    \begin{tabular}{lll}
        \toprule
        Method & $\map$ & $\cmap$\\
        \midrule
        Source & 63.8 & 46.9 \\
        AdaBN & 53.0 & 38.5 \\
        SHOT & 67.5 & 50.3 \\
        TENT & 53.5 & 38.6 \\
        PL & 72.0 & 51.1 \\
        DS & 76.8 & 52.7 \\
        DUST & 76.0 & 51.6 \\
        NRC & 66.1 & 49.3 \\
        \rowcolor{lightgreen!25} \ourmethod{} & {\bf 78.1} & {\bf 54.4} \\
        \bottomrule
    \end{tabular}
    \label{tab:validation}
\end{table}

{\bf Results} We found neither AdaBN nor TENT were able to improve the source model, regardless of the hyperparameter configuration or the domain (\autoref{tab:bioacoustics}; also see \autoref{fig:test_plots} for a hindsight analysis). This is surprising for TENT, since EM has also previously shown strong performance on various tasks~\citep{wang2021tent,vu2019advent}. We hypothesize this is due to a combination of extreme distribution shift as well as the multi-label nature of our problem. Indeed, in the single-label scenario minimizing entropy leads to forcing the model to choose a single class for each example in an increasingly confident manner. On the other hand, in our multi-label scenario there is no constraint that a class should be chosen. This fact---combined with the low confidence caused by very large distribution shifts (\autoref{fig:model_calibration}, (1,2))---can drive the model to a collapsed state where all class probabilities are zero (\autoref{fig:model_calibration}, (3)).

As for SHOT and PL, we were able to find hyper-parameters that improved over the source model's performance on the validation set, as observed in \autoref{tab:validation}. We observed however that the gains did not consistently translate to the test domains (\autoref{tab:bioacoustics}). For example, PL boosted the source model by~$8.2\%$ $\map$ on the validation domain, while degrading it by more than $23\%$ $\map$ on the Hawai'i test domain. 

On the other hand, we find that DTS works significantly better on our challenging shifts, but suffers from stability issues: it often displays a plateau followed by a degradation of the model's performance (\autoref{fig:test_plots}; see e.g. Powdermill). NRC and DUST are also able to improve upon the source model in some cases, but are not consistent in achieving this for all of the shifts we study, and also suffer from the same plateau-and-degradation trend (e.g. see Hawai'i for NRC and Powdermill for DUST). This trend would necessitate a precise early-stopping procedure to pinpoint the number of adaptation updates to perform before degradation starts. This is a serious drawback in the context of SFDA, given the absence of a domain-specific labelled validation set for tuning the training schedule or performing early-stopping.

In contrast, we find that \ourmethod{}~not only addresses these stability issues, but also outperforms all considered baselines, setting the state-of-the-art on the bioacoustics shifts. 

\begin{table}[t]
    \caption{Validation and ablation results from the High Sierras bioacoustics dataset. Dropout Noise, Softness and Laplacian regularization ingredients \textbf{act symbiotically} to provide the best performances. Removing any ingredient leads to significant drops in performances.}
    \vskip 0.15in
    \centering
    \footnotesize
    \begin{tabular}{ccccc}
    \toprule
    Dropout & Softness & Laplace Reg.& mAP & cmAP \\
    \midrule
    \multicolumn{3}{c}{\em pre-trained model (no adaptation) } & 63.8 & 46.9 \\
     & \cmark & \cmark & 61.3 & 33.3 \\
    \cmark &  & \cmark & 72.1 & 49.4 \\
    \cmark & \cmark &  & 77.0 & 52.0 \\
    \cmark & \cmark & \cmark & {\bf 78.1} & {\bf 54.4} \\
    \bottomrule
    \end{tabular}
    \label{tab:ablation}
\end{table}

{\bf Disentangling the multi-label effect}  We previously conjectured that the multi-label nature of the bio-acoustics tasks contributed to SFDA methods' failure. We challenge that supposition by generating a new version of each bio-acoustics dataset, in which all recordings containing more than a single bird annotation are filtered. That allows us to fall back onto the standard single-label setting, in which the model's logits can be constrained into a single probability distribution through a softmax operator. Results are presented in \autoref{tab:single_label_bioacoustics}. While an exact apple-to-apples comparison with the multi-label case is not possible, because a lot of recordings have been filtered, we can already draw interesting insights. First, the single-label constraint (through the use of softmax) appears to be a strong regularizer that can substantially contribute to the success of SFDA methods. For instance, the pseudo-labelling baseline (PL), which dramatically failed on 2 test cases in the multi-label scenario, now systematically outperforms the baseline, and even reaches a state-of-the art performance on Powdermill. However, \autoref{tab:single_label_bioacoustics} also confirms our premise that the multi-label factor cannot alone fully explain methods failing in the audio domain, as TENT and AdaBN still suffer from dramatic model collapses on several tests domains. Similarly, SHOT which obtains strong results on single-label visual tasks, as shown in \autoref{sec:visual_task}, performs poorly in this setting. Overall, \ourmethod{} preserves its ability to systematically improve over the baseline, and obtains state-of-the-art results on 4 out of 6 target domains.

{\bf Ablations on \ourmethod{}} We find in \autoref{tab:ablation} that the presence of the noise is crucial, and that removing it leads to worse performance than the non-adapted model. Furthermore, removing softness in labels ($\alpha=0$) or Laplacian Regularization ($\lambda=0$) significantly under-performs the full method, thereby highlighting the symbiosis of all three components.

\begin{table*}[t]
    \caption{Top-1 accuracy (averaged over 5 random seeds) on vision test benchmarks. \ourmethod{}~approaches best methods on CIFAR-10-C, and surpasses them on ImageNet variants. We report per-corruption confidence-intervals for CIFAR-10-C in \autoref{sec:additional_results}}
    \label{tab:vision}
    \vskip 0.15in
    \begin{center}
    \footnotesize
    \begin{tabular}{lcccc}
    \toprule
    Method                       & \begin{tabular}{c}CIFAR-10-C \\ {\scriptsize(Average across severity 5 corruptions)} \end{tabular} & ImageNet-R & ImageNet-Sketch & Vis-DA-C \\
    \midrule
    Source                       & $56.66$ & $23.16 \pm 0.0$ & $21.67 \pm 0.0$ & $51.8 \pm 0.0$ \\
    AdaBN~\citep{li2018adaptive} & $80.10$ & $24.54 \pm 0.08$ & $22.55 \pm 0.14$ & $65.63 \pm 0.3$ \\
    SHOT~\citep{liang2020we}     & $82.63$ & $24.49 \pm 0.14$ & $26.16 \pm 0.1$ & $\textbf{76.06} \pm 0.14$ \\
    TENT~\citep{wang2021tent}    & $82.69$ & $28.78 \pm 0.04$ & $27.56 \pm 0.16$ & $71.03 \pm 0.25$ \\
    PL~\citep{lee2013pseudo}     & $\bf 83.85$ & $26.54 \pm 3.43$ & $2.03 \pm 1.5$ & $50.16 \pm 0.67$ \\
    NRC~\cite{yang2021exploiting}   & $80.58$ & $25.73 \pm 0.1$ & $23.66 \pm 0.1$ & $69.13 \pm 0.22$ \\
    DUST~\citep{khurana2021unsupervised}    & $80.07$ & $24.61 \pm 0.09$ & $22.54 \pm 0.07$ & $65.29 \pm 0.21$ \\
    Dropout Student              & $82.18$ & $29.14 \pm 0.14$ & $27.49 \pm 0.11$ & $69.9 \pm 0.18$ \\
    \rowcolor{lightgreen!25} \ourmethod{}~(ours) &  $82.21$ & $\textbf{32.97} \pm 0.11$ & $\textbf{30.92} \pm 0.1$ & $72.08 \pm 0.33$ \\
    \bottomrule
    \end{tabular}
    \end{center}
\end{table*}
    
\subsection{Vision Tasks} \label{sec:visual_task}

While we have shown that recent SFDA approaches are not as generalizable as previously thought, we have yet to show that the simple \ourmethod~approach developed in the context of bioacoustics distribution shifts is generalizable enough to perform well on vision tasks.

\textbf{Data processing and source models.}  We process vision datasets in accordance with their respective established practices. We adopt model architectures from previous works, namely a ResNet-50~\citep{he2016deep} for ImageNet benchmarks, a Wide ResNet 28-10~\citep{zagoruyko2016wide} for CIFAR-10 benchmarks, and a ResNet-101 for VisDA-C. We use the same CIFAR-10 Wide ResNet model checkpoint~\citep[provided by][]{croce2021robustbench} as \citet{wang2021tent} and the ResNet-101 checkpoint provided by \citet{yang2021exploiting} for fair comparison. 

{\bf Metrics and hyperparameter selection.} We report top-1 accuracy for all vision SFDA benchmarks. We use the three most challenging corruptions from ImageNet-C (\textit{contrast, glass blur and snow}) as the validation domain for all vision tasks (we use the average accuracy across those three).
        
\textbf{Results.} From \autoref{tab:vision}, we find that \ourmethod{} consistently exhibits strong performance on vision benchmarks, despite the change in modality and the single-label setup. It performs strongly on all datasets and in fact outperforms all baselines on both ImageNet-R and ImageNet-Sketch.

Interestingly, SHOT performs uncharacteristically well on VisDA-C, whereas NRC underperforms SHOT, TENT, and \ourmethod. We note that several factors could explain this observation. We purposefully split VisDA-C's target domain into an adaptation and an evaluation set, unlike the TTA procedure used in the NRC paper. We also performed a full hyperparameter search for NRC (as well as all other approaches considered) using ImageNet-C as a validation domain, as described earlier, which naturally may yield less optimistic results compared to performing model selection on the target domain. In line with our overall takeaway, this is possibly indicative of a lack of performance consistency across different evaluation methodologies.

\section{Discussion and Conclusion}

We investigated the generalizability of recent source-free domain adaptation methods developed in the context of vision tasks. We observe that when applied to a new modality and problem setting (multi-label classification of audio recordings of wild bird vocalizations), these methods' performance characteristics differ greatly from observations made when evaluating on vision tasks (and in some cases fail to live up to expectations set in the vision domain).

In light of this, our first message is that as more and more SFDA approaches are developed for vision tasks, they may become increasingly co-adapted to the characteristics of that domain, up to a point where progress there ceases to translate into progress on other modalities and problems. Which characteristics the methods co-adapt to is not yet clear, as the audio classification task considered in this work differs in many ways from the image classification tasks used as standardized SFDA benchmarks. Beyond the shift from vision to audio modality, and from a single-label to multi-label problem, other confounders could be responsible for our observations, such as the severe class imbalance, the extremity of the distribution shifts and the complex nature of bird vocalizations that have distinct modes (songs vs. calls) even within the same class.

Our work intends to draw attention to the surprising lack of generalizability of SFDA methods and encourage practitioners to expand the scope of their evaluation (for instance using our proposed bird song classification task). We also hope to encourage future work into properly characterizing which of the above factors contribute to the observed differences in performance characteristics, in order to gain a deeper understanding into existing SFDA methods and on how to improve them. 

Our second message is that consistent and generalizable performance is a valuable attribute for SFDA approaches given the nature of the problem (no target labels, no well-defined validation set) and the resulting challenge in performing model selection. One may spend a lot of time and effort developing a strong approach for a particular pair of source and target domains only to see it underperform in a different context. Mitigating this danger requires careful research methodology, isolating whole development datasets for hyper-parameter tuning. From this perspective, an approach which performs consistently well in multiple contexts (even if sometimes worse than top-performing approaches in each individual domains) is valuable in practice. While we don't believe that \ourmethod~ is the ultimate SFDA method, we believe it is a strong baseline in terms of this consistency desideratum while performing very competitively: it sets the state-of-the-art on our challenging bioacoustics shifts and is a strong competitor on vision benchmarks.

Moreover, we argue that beyond evaluation, our proposed bioacoustics task is important for model development in and of itself, as it surfaces differences in performance characteristics which, when addressed in our proposed \ourmethod~approach, resulted in the desired generalizable performance.

\section*{Acknowledgements}

Each author of this paper contributed in the following way:
\begin{itemize}
    \item Malik proposed the method presented in this work. He implemented it and all methods we compare against, conducted all experiments, and produced all figures and tables.
    \item Tom implemented several soundscapes datasets used by Malik for his experiments and was a significant contributor to the codebase used by Malik for his experiments.
    \item Bart implemented the bioacoustics classifier and was a significant contributor to the codebase used by Malik for his experiments.
    \item Vincent co-advised Malik on the project. He contributed the Xeno-Canto dataset implementation and was a significant contributor to the codebase used by Malik for his experiments.
    \item Eleni co-advised Malik on the project. She influenced the project direction and contributed to the framing of the proposed approach.
\end{itemize}

\newpage
\bibliography{notela_icml_2023}
\bibliographystyle{icml2023}

\newpage
\appendix
\onecolumn
\section{Bioacoustics Datasets}
\label{sec:datasets_long}

\begin{table*}[t]
    \caption{\textbf{Relationship of problem settings}. $x$ and $y$ denote inputs and labels, and $s$ and $t$ ``source'' and ``target'', respectively (note that in some cases, as in DG, $s$ might be a union of source domains / environments). For TTA and SFDA, the $*$ in their training data and loss reflects that they are entirely agnostic to how source training is performed, allowing the use of any off-the-shelf model.}
    \label{tab:problem_settings}
    \vskip 0.15in
    \begin{center}
    \footnotesize
    \begin{tabular}{llllll}
    \toprule
                                        & DA                & DG            & TTT           & TTA           & SFDA   \\
    \midrule
    Data used for training              & $x^s,y^s,x^t$     & $x^s,y^s$     & $x^s,y^s$     & $*$           & $*$    \\
    Data used for adaptation            & $x^s,y^s,x^t$     & $x^s,y^s,x^t$ & $x^t$         & $x^t$         & $x^t$  \\
    \midrule
    Training loss                       & $\mathcal{L}(x^s,y^s) + \mathcal{L}(x^s, x^t)$     & $\mathcal{L}(x^s,y^s)$   & $\mathcal{L}(x^s,y^s) + \mathcal{L}(x^t)$     & $*$           & $*$    \\
    Adaptation loss                     & ---                                                & ---                      & $\mathcal{L}(x^t)$         & $\mathcal{L}(x^t)$         & $\mathcal{L}(x^t)$  \\
        \midrule
         \\
        \end{tabular}
        \end{center}
\end{table*}

We use {\bf Xeno-Canto}~\citep[XC;][]{vellinga2015xeno} as the source dataset for bird species classification in the audio domain. XC is a growing, user-contributed collection of Creative Commons recordings of wild birds across the world. Our snapshot, downloaded on July 18, 2022, contains around 675,000 files spanning 10,932 bird species. Recordings are {\em focal} (purposefully capturing an individual's vocalizations in natural conditions, as opposed to {\em passively} capturing all ambient sounds), and each is annotated with a single foreground label (for the recording's main subject) and optionally a varying number of background labels (for other species vocalizing in the background).

For our distributionally-shifted datasets, we use multiple collections of passive (also called {\em soundscape}) recordings from various geographical locations. We use a 75/25 \% split to obtain $\sD_t^{adapt}$---used to adapt the model---and $\sD_t^{test}$---used to evaluate the adapted model.

\begin{itemize}
    \item {\bf Sapsucker Woods}~\citep[SSW;][]{kahl2022ssw} contains soundscape recordings from the Sapsucker Woods bird sanctuary in Ithaca, NY, USA.
    \item {\bf Sierra Nevada}~\citep{kahl2022sierra} contains soundscape recordings from the Sierra Nevada in California, USA.
    \item {\bf Hawai'i}~\citep{navine2022hawaii} contains soundscape recordings from Hawai'i, USA. Some species, particularly endangered honeycreepers, are endemic to Hawai'i and many are under-represented in the Xeno-Canto training set.
    \item {\bf Powdermill} (\cite{chronister2021annotated}) contains high-activity dawn chorus recordings captured over four days in Pennsylvania, USA.
    \item {\bf Caples} is an unreleased dataset collected by the California Academy of Science at the Caples Creek area in the central Californian Sierra Nevadas. Work is underway to open-source this dataset.
    \item {\bf Colombia} is an unreleased dataset, previously used as part of the test set for the BirdCLEF 2019 competition.
    \item {\bf High Sierras} is an unreleased dataset, previously used as part of the test set for the Kaggle Cornell Birdcall Identification challenge. Recordings are typically sparse, but with very low SNR due to wind noise. Work is underway to open-source this dataset.
\end{itemize}

\section{Xeno-Canto data processing}
\label{app:xc_processing}

Xeno-Canto recordings range from less than 1 second to several hours long. To extract 6-second segments we use a heuristic to identify segments with strong signal.

\begin{enumerate}
    \item If the audio is shorter than 6 seconds, pad the recording evenly left and right using wrap-around padding.
    \item Convert the audio into a log mel-spectrogram.
    \item Denoise the spectrogram:
    \begin{enumerate}
        \item For each channel calculate the mean and standard deviation. All values that lie more than 1.5 standard deviations from the mean are considered outliers.
        \item Calculate a robust mean and standard deviation using the inliers\footnote{In the calculation of the mean and variance we add 1 to the denominator to avoid division by zero in the case that all values are considered outliers.}.
        \item Any values that lie more than 0.75 robust standard deviations away from the robust mean are considered signal. Shift the signal in each channel by its robust mean, and set all noise to zero.
    \end{enumerate} 
    \item Sum all channels in the denoised spectrogram to create a signal vector.
    \item Use SciPy's \texttt{find\_peaks\_cwt} function to retrieve peaks, using 10 Ricker wavelets with widths linearly spaced between 0.5 and 2 seconds
    \item Select windows of 0.6 seconds centred at each peak and discard the peak if the maximum value in this window is smaller than 1.5 times the mean of the signal vector.
    \item Keep only up to 5 peaks, with the highest corresponding values in the signal vector.
    \item Select a 6 second window centred at each peak. If the window overlaps the start or beginning of the boundary, shift the window accordingly.
\end{enumerate}

\begin{table}[t]
    \caption{Summary of Bioacoustic Soundscape Dataset Characteristics. \\ `XC/Species' indicates the average number of Xeno-Canto training example files per species. `Low Data Species' are species with fewer than 50 training examples available. Labels per example is computed on the peak-sliced data, while hours and number of labels refer to the original raw dataset.}
    \label{tab:biodata}
    \begin{center}
    \footnotesize
    \centering
    \begin{tabular}{l >{\centering\arraybackslash}b{1cm} >{\centering\arraybackslash}b{1cm} >{\centering\arraybackslash}b{1cm} >{\centering\arraybackslash}b{1.15cm} >{\centering\arraybackslash}b{1.3cm}  | >{\centering\arraybackslash}b{1.15cm} >{\centering\arraybackslash}b{1.15cm}}
    \toprule
    & Hours &  \#Species & \#Labels & Labels/ Example & Climate & XC/ Species & Low data \\
    \midrule
    Sapsucker        & 285 & 96 & 50,760  & 1.5 & Temperate & 367.9 & 3\% \\
    Sierra Nevada   & 33 & 56 & 20,147   & 1.9 & Temperate & 416.0 & 0\% \\
    Hawai'i         & 51 & 27 & 59,583   & 1.5 & Tropical & 166.3 & 44\% \\ 
    Powdermill      & 6 & 48 & 16,052    & 3.2 & Temperate & 360.0 & 0\% \\
    Caples          & 6 & 78 & 4,993     & 1.4 & Temperate & 334.8 & 10\% \\
    Colombia        & 8 & 63 & 1,489     & 1.4 & Tropical & 215.8 & 6\% \\
    High Sierras    & 34 & 19 & 14,494   & 1.2 & Alpine & 323.5 & 5\% \\
    \bottomrule
    \end{tabular}
    \end{center}
\end{table}

\section{Soundscapes data processing}
\label{app:soundscapes_processing}

We extract 5-second segments from soundscapes recording by cross-referencing the bounding box labels with the same heuristic used to extract 6-second segments from XC recordings:

\begin{enumerate}
    \item Use the procedure outlined in Appendix~\ref{app:xc_processing} to extract 5-second (rather than 6-second) windows from up to 200 (rather than 5) high-signal peaks per source file.
    \item For all 5-second windows:
    \begin{enumerate}
        \item If it does not overlap in time with any bounding box label, drop it.
        \item Otherwise, find all overlapping bounding box labels and label the window with the union of all their labels.
    \end{enumerate}
\end{enumerate}

\section{Metrics}
\label{sec:metrics_equations}

\subsection{mAP}
        
Define $\precision_X(s, c)$ as the generalized inverse rank of a ground-truth positive label $c$ in an observation $s$, computed as:
\[
  \precision_X(s, c) = \frac{1}{\rank_X(s, c)} \sum_{r=1}^{\rank_X(s,c)} \mathbf{1}[\lab(s, c) \in \mathcal{C}(s)]
\]
where $X$ is the corpus over which we perform rankings, $\rank_X{s, c}$ is the rank of the score for class $c$ in observation $s$ in the corpus $X$, $\lab(s, r)$ is the ground-truth label of class $r$ in observation $s$, and $\mathcal{C}(s)$ is the set of ground-truth positive classes in observation $s$. Finally, $\mathbf{1}[\lab(s, r) \in \mathcal{C}(s)]$ is the indicator function for whether observation $s$ is a ground-truth member of class $c$.

The $\map$ metric measures the per-example precision, averaged over the set $\mathcal{E}$ of all examples in the dataset:
\[
    \frac{1}{|\mathcal{E}|} \sum_{s\in \mathcal{E}} \frac{1}{|\mathcal{C}(s)|} \sum_{c\in \mathcal{C}(s)} \precision_{\mathcal{E}}(s, c)
\] 
            
\subsection{cmAP}

Class-wise mean average precision ($\cmap$) is defined as:
\[
    \frac{1}{|\mathcal{C}|} \sum_{c\in \mathcal{C}} \frac{1}{|c \in \mathcal{E}|} \sum_{s\in \mathcal{E}} \precision_{\mathcal{C}}(s, c)
\]
Here $|c \in \mathcal{E}|$ denotes the total number of ground-truth positive examples of class $c$ in the dataset. Notice that in this case, the precision ranking is over the logits for each target class, instead of ranking the logits in a single observation.
    
\section{Hyperparameter validation}

As mentioned in \autoref{sec:results}, we reproduce all methods and ensure fairness of comparisons by (i) using the same pre-trained models and (ii) re-tuning each method's hyperparameters. All experiments are carried out with a batch size set to 64 (both audio and vision). Table \ref{tab:hyperparameters} displays the grids used for hyper-parameter tuning (also both for audio and vision tasks). Note that to reduce the load of hyperparameters to tune in \ourmethod{}, we make the design choice of using $\lambda=\alpha$, which effectively removes one degree of freedom.

\begin{table}[t]
    \caption{Grid used for tuning hyper-parameters.}
    \label{tab:hyperparameters}
    \centering
    \begin{tabular}{lll}
         \toprule
         Method & Hyperparameter & Grid \\
         \midrule
         \multirow{5}{*}{All} & Learning rate & \{ 1e-5, 1e-4, 1e-3 \} \\
          & Trainable parameters & \{ BatchNorm scale and bias, all \} \\
           & Use of dropout & \{True, False \} \\
          & Use source BN statistics & \{True, False \} \\
          & Learning rate cosine decay & \{True, False \} \\
         \midrule
         SHOT \citep{liang2020we} & Pseudo-labels weight $\beta$ & \{0., 0.3, 0.6, 0.9\} \\
         \midrule
         Pseudo-labelling \cite{lee2013pseudo} & Confidence threshold & \{0., 0.5, 0.9, 0.95\} \\
         \midrule
         \multirow{2}{*}{Dropout Student} & Softness weight $\alpha$ & \{0.1, 1.0\} \\
                                          & Pseudo-label update frequency & \{Every iteration, Every epoch\} \\
         \midrule
         \multirow{2}{*}{DUST} & Number of random passes & \{2, 3, 4\} \\
                                          & KL threshold & \{0.8, 0.9, 0.99\} \\
         \midrule
         \multirow{2}{*}{NRC} & $k$ nearest neighbors & \{5, 10, 15\} \\
                              & $k$ extended nearest neighbors & \{5, 10, 15\} \\
                              & Base affinity & \{0.1, 0.2\} \\
         \midrule
         \multirow{3}{*}{NOTELA} & $k$ nearest neighbors & \{5, 10, 15\} \\
                                 & Softness weight $\alpha$ & \{0.1, 1.0\} \\
                                 & Pseudo-label update frequency & \{Every iteration, Every epoch\} \\
          \bottomrule
    \end{tabular}
\end{table}

\section{Proof of \autoref{eq:laplacian_updates}} \label{sec:laplacian_proof}

We hereby provide the proof, as well as a more formal justification for the updates given in \autoref{eq:laplacian_updates}. We start by restating the objective we want to minize:
\begin{align}  \label{eq:proof_11}
    \min_{\ybf_{1:N}}& \quad \Tr\left(- \frac{1}{N} \sum_{i=1}^N \ybf_i^\top \log(\pbf_i) + \frac{\alpha}{N} \sum_{i=1}^N \ybf_i^\top \log(\ybf_i) - \frac{\lambda}{N} \sum_{i=1}^N \sum_{j=1}^N w_{ij} ~ \ybf_i^\top \ybf_j\right) \nonumber \\
    \text{s.t}& \quad \mathbf{1}^\top\ybf_i  = 1, ~ \ybf_i \geq 0\,.
\end{align}
\textbf{General case.} Let us consider an easier problem, in which the Laplacian term has been linearized around the current solution $\ybf_i=\pbf_i$,
\begin{align} \label{eq:proof_12}
    \min_{\ybf_{1:N}}& \quad \Tr\left(- \frac{1}{N} \sum_{i=1}^N \ybf_i^\top \log(\pbf_i) + \frac{\alpha}{N} \sum_{i=1}^N \ybf_i^\top \log(\ybf_i) - \frac{\lambda}{N} \sum_{i=1}^N \sum_{j=1}^N w_{ij} ~ \ybf_i^\top \pbf_j \nonumber \right) \\
    \text{s.t}& \quad \mathbf{1}^\top\ybf_i  = 1\,.
\end{align}
We purposefully omitted the $ \ybf_i \geq 0$ constraint, which will be satisfied later on by our solution to \autoref{eq:proof_11}. The Lagrangian of this problem is
\begin{align}
    \mathcal{L}(\ybf_{1:N}) = &\Tr\left(- \frac{1}{N} \sum_{i=1}^N \ybf_i^\top \log(\pbf_i) + \frac{\alpha}{N} \sum_{i=1}^N \ybf_i^\top \log(\ybf_i) - \frac{\lambda}{N} \sum_{i=1}^N \sum_{j=1}^N w_{ij} ~ \ybf_i^\top \pbf_j   \right) \\
    &+ \frac{1}{N} \sum_{i=1}^N \gamma_i (\bf 1^\top \ybf_i - 1)\,,
\end{align}
and the gradient of this Lagrangian with respect to $\ybf_i$ is
\begin{align}
    N.\nabla_{\ybf_i} \mathcal{L} = - \log(\pbf_i) + \alpha(\log(\ybf_i) + \mathbf{1}) - \lambda \sum_{j=1}^N w_{ij} \pbf_j + \gamma_i \mathbf{1}\,.
\end{align}
Solving for $\ybf_i$ yields
\begin{align}
    \ybf_i = \exp \left(-\frac{\alpha + \gamma_i}{\alpha}\right) \exp\left(\frac{\lambda}{\alpha}\sum_{j=1}^N w_{ij} \pbf_j \right) \odot \pbf_i^{1/\alpha}\,.
\end{align}
Now $\gamma_i$ is chosen such that the constraint $\ybf_i^\top \mathbf{1} = 1$ is satisfied, resulting in
\begin{align}
    \ybf_i \propto \pbf_i^{1/\alpha} \odot \exp\left(\frac{\lambda}{\alpha}\sum_{j=1}^N w_{ij} \pbf_j \right)\,.
\end{align}

\begin{table}[t]
    \centering
    \caption{Top-1 accuracy, averaged over 5 random seeds, along with $95 \%$ confidence interval for each corruption in CIFAR-10-C.}
    \label{tab:full_cifar}
    \resizebox{\textwidth}{!}{
\begin{tabular}{llllllllll}
\toprule
method &             Baseline &         AdaBN &              DUST &   Dropout Student &            NOTELA &               NRC &           Pseudo-Labelling &                       SHOT &              TENT \\
corruption &                   &                  &                   &                   &                   &                   &                            &                            &                   \\
\midrule
brightness &   $91.28 \pm 0.0$ &  $92.2 \pm 0.03$ &  $92.2 \pm 0.0$ &  $92.43 \pm 0.16$ &   $92.53 \pm 0.1$ &  $92.35 \pm 0.05$ &           $92.49 \pm 0.24$ &  $\textbf{92.68} \pm 0.16$ &   $92.5 \pm 0.09$ \\
contrast   &   $52.84 \pm 0.0$ &  $87.3 \pm 0.02$ & $87.02 \pm 0.13$ &  $88.89 \pm 0.36$ &  $89.75 \pm 0.23$ &  $87.55 \pm 0.11$ &  $\textbf{89.94} \pm 1.63$ &           $89.21 \pm 0.12$ &  $89.08 \pm 0.32$ \\
defocus    &   $53.16 \pm 0.0$ &  $87.4 \pm 0.09$ & $87.35 \pm 0.07$ &  $88.42 \pm 0.24$ &  $88.47 \pm 0.22$ &  $87.61 \pm 0.09$ &  $\textbf{89.01} \pm 0.43$ &           $88.45 \pm 0.13$ &  $88.68 \pm 0.07$ \\
elastic    &  $73.28 \pm 0.0$  &  $77.02 \pm 0.1$ &  $77.09 \pm 0.05$ &   $78.6 \pm 0.39$ &   $78.4 \pm 0.22$ &  $77.55 \pm 0.07$ &   $\textbf{80.2} \pm 0.39$ &            $79.07 \pm 0.1$ &  $79.02 \pm 0.07$ \\
fog        &  $74.08 \pm 0.0$  &  $85.92 \pm 0.04$ & $85.91 \pm 0.04$ &   $87.3 \pm 0.26$ &  $86.94 \pm 0.38$ &  $85.99 \pm 0.05$ &  $\textbf{88.49} \pm 0.33$ &           $87.22 \pm 0.16$ &  $87.46 \pm 0.13$ \\
frost      &  $59.6 \pm 0.0$   & $82.45 \pm 0.05$ &  $82.46 \pm 0.07$ &  $84.47 \pm 0.37$ &   $84.55 \pm 0.3$ &  $83.26 \pm 0.04$ &   $\textbf{85.34} \pm 0.7$ &           $84.71 \pm 0.15$ &  $84.86 \pm 0.06$ \\
frosted    &  $46.36 \pm 0.0$  & $65.38 \pm 0.03$ &  $65.26 \pm 0.1$ &  $69.38 \pm 0.43$ &  $69.05 \pm 0.11$ &   $66.07 \pm 0.1$ &    $\textbf{71.7} \pm 0.5$ &           $70.58 \pm 0.21$ &  $70.08 \pm 0.15$ \\
gaussian   &  $28.16 \pm 0.0$  & $73.31 \pm 0.03$ &  $73.26 \pm 0.07$ &  $75.75 \pm 0.36$ &  $75.44 \pm 0.18$ &  $73.93 \pm 0.08$ &  $\textbf{78.62} \pm 0.53$ &           $76.62 \pm 0.08$ &  $76.58 \pm 0.13$ \\
impulse    &  $28.04 \pm 0.0$  &  $64.36 \pm 0.05$ & $64.34 \pm 0.06$ &   $67.74 \pm 0.5$ &  $67.82 \pm 0.32$ &  $65.45 \pm 0.08$ &  $\textbf{70.86} \pm 1.05$ &           $68.86 \pm 0.11$ &  $68.96 \pm 0.17$ \\
jpeg       &  $69.12 \pm 0.0$ &  $72.57 \pm 0.06$ & $72.49 \pm 0.06$ &   $75.57 \pm 0.2$ &  $75.32 \pm 0.14$ &  $73.14 \pm 0.09$ &  $\textbf{79.71} \pm 0.27$ &           $76.01 \pm 0.17$ &  $76.24 \pm 0.21$ \\
motion     &   $65.16 \pm 0.0$ &  $86.29 \pm 0.05$ & $86.33 \pm 0.07$ &  $87.72 \pm 0.09$ &  $87.96 \pm 0.12$ &  $86.51 \pm 0.05$ &   $\textbf{88.12} \pm 0.3$ &           $87.98 \pm 0.12$ &    $87.9 \pm 0.1$ \\
pixelate   &  $41.84 \pm 0.0$  &  $80.7 \pm 0.04$ & $80.66 \pm 0.03$ &    $83.5 \pm 0.2$ &  $83.66 \pm 0.29$ &   $81.3 \pm 0.08$ &  $\textbf{85.71} \pm 0.66$ &           $83.66 \pm 0.15$ &   $84.1 \pm 0.11$ \\
shot       &  $34.52 \pm 0.0$  &  $75.68 \pm 0.07$ &  $75.7 \pm 0.05$ &  $78.64 \pm 0.21$ &  $78.58 \pm 0.35$ &  $76.36 \pm 0.04$ &   $\textbf{81.42} \pm 0.4$ &           $79.54 \pm 0.15$ &  $79.88 \pm 0.24$ \\
snow       &  $75.12 \pm 0.0$  &  $82.88 \pm 0.03$ & $82.91 \pm 0.05$ &   $84.98 \pm 0.2$ &  $85.17 \pm 0.17$ &  $83.18 \pm 0.06$ &  $\textbf{85.92} \pm 0.36$ &           $85.22 \pm 0.15$ &  $85.35 \pm 0.15$ \\
zoom       &  $57.36 \pm 0.0$ &  $88.02 \pm 0.05$ &  $87.99 \pm 0.04$ &  $89.25 \pm 0.43$ &    $89.5 \pm 0.2$ &  $88.41 \pm 0.09$ &   $\textbf{90.1} \pm 0.37$ &           $89.62 \pm 0.16$ &  $89.64 \pm 0.14$ \\
\bottomrule
\end{tabular}
}
\end{table}

\paragraph{Concavity.} Let $\mathbf{W} = (w_{ij}) \in \mathbb{R}^{N\times N}$ be the matrix of affinity weights. An additional assumption on $\bf{W}$ allows a more formal justification of the linearization of the Laplacian term. Specifically, we can justify that if $\bf W + \bf W^\top$ is positive semi-definite, then the last term in \autoref{eq:proof_11} is concave.

To show this rewrite $\Tr\left(\sum_{i=1}^N \sum_{j=1}^N w_{ij} ~ \ybf_i^\top \ybf_j \right)$ as $\vone^\top (\mW \odot \mY^\top \mY)\vone$ where $\mY = (\mathrm{vect}(\ybf_i))$, which is in $\mathbb{R}^{N\times C}$ or $\mathbb{R}^{N\times2C}$ for the single- and multi-label case respectively. The Hessian of this function is $\mW\otimes \mathbb{I}+\mW^\top \otimes \mathbb{I}$, whose eigenvalues are multiplicities of those of $\mW + \mW^\top$.

Hence equation \ref{eq:proof_11} can be solved by a concave-convex procedure \citep[CCP;][]{yuille2003concave, ziko2018scalable, boudiaf2022parameter}, which is suited to cases in which one part of the objective is convex (the two first terms in our case) and the other is concave (the Laplacian term). CCP proceeds by minimizing a sequence of \textit{pseudo-bounds}, i.e., an upper bound that is tight at the current solution, obtained by linearizing the concave part of the objective at the current solution. Therefore, our proposed updates from \autoref{eq:laplacian_updates} can be interpreted as the first iteration of this procedure. Starting from the initial solution, $\ybf_i ^{(0)}=\pbf_i$, unrolling the CCP procedure would consist of performing the following updates until convergence:
\begin{align}
    \ybf_{i}^{(t)} \propto \pbf_i^{1/\alpha} \odot \exp\left(\frac{\lambda}{\alpha}\sum_{j=1}^N w_{ij} \ybf_j^{(t-1)}\right)\,.
\end{align}
    

\paragraph{Affinity weights.} Let $\mA$ be the adjacency matrix of the mutual $k$-nearest neighbours graph, and $\mD$ the diagonal matrix with the node degrees (i.e., $(\mD)_{ii}$ is the number of mututal neighbours for sample $i$). We know that $\mA +\mD$ is positive semi-definite~\citep{desai1994characterization} and hence matrix $\mW = \mD^{-\frac{1}{2}}(\mA + \mD)\mD^{-\frac{1}{2}}$ is also positive semi-definite. Note that this is equivalent to the case where $w_{ij}$ is set to the reciprocal of the number of mutual neighbours, and $w_{ii} = 1$.

The terms $w_{ii}$ act as an L2 regularizer on the values $\ybf_i$. Note that in our experiments we deviate from the theory and set $w_{ii} = 0$ to have better control over the regularization of $\ybf_i$.
    
\section{Additional results}
\label{sec:additional_results}
    

\paragraph{CIFAR-10-C per-corruption results.} We provide the per-corruption results on CIFAR-10-C in \autoref{tab:full_cifar}, as well as the  $95 \%$ confidence intervals.



\end{document}